\documentclass{article} 

\usepackage{iclr2019_conference}

\usepackage{times}
\usepackage{latexsym}

\usepackage[utf8]{inputenc} 
\usepackage[T1]{fontenc}    
\usepackage{hyperref}       
\usepackage{url}            
\usepackage{booktabs}       
\usepackage{amsfonts}       
\usepackage{nicefrac}       
\usepackage{microtype}      
\usepackage{lipsum}
\usepackage{graphicx}

\iclrfinalcopy

\title{EpiRL: A Reinforcement Learning Agent \\ to Facilitate Epistasis Detection}

\author{Kexin Huang\\
Courant Institute\\New York University\\
521 Mercer Street, New York, NY 10012\\
\texttt{kexin.huang@nyu.edu} \\
\And
Rodrigo Nogueira\\
Tandon School of Engineering\\New York University\\
6 MetroTech Center, Brooklyn, NY 11201\\
\texttt{rodrigonogueira@nyu.edu}
}

\begin{document}
\maketitle

\begin{abstract}
Epistasis (gene-gene interaction) is crucial to predicting genetic disease. Our work tackles the computational challenges faced by previous works in epistasis detection by modeling it as a one-step Markov Decision Process where the state is genome data, the actions are the interacted genes, and the reward is an interaction measurement for the selected actions. A reinforcement learning agent using policy gradient method then learns to discover a set of highly interacted genes.
\end{abstract}


\section{Introduction}
The fundamental goal for studying genetics is to understand how certain genes can incur disease and traits. Since the advent of Genome-Wide Association Studies (GWAS)~\citep{burton2007genome}, thousands of SNP (Single Nucleotide Polymorphism)s have been identified and associated with genetic diseases and traits. These SNPs are discovered through one-SNP-at-a-time statistical analysis. However, individual gene marker is insufficient to explain many complex diseases and traits~\citep{mackay2014epistasis}. Now, most geneticists believe that gene-gene interaction (epistasis) can explain the missing heritability incurred by the traditional approach. 

There has been a substantial amount of work on epistasis detection. Exhaustive combinatorial search methods like Multifactor Dimensionality Reduction (MDR)~\citep{yang2018multiple} have been shown successful, but only in small genome-scale due to computational complexity. Later, attempts to reduce search spaces exhibit efficiency, like ReliefF and Spatially Uniform ReliefF~\citep{niel2015survey}. Besides, machine learning-based algorithms gain popularity. Most of the machine learning based methods for epistasis detection model the epistatic process as a non-linear neural network. It predicts if an input sequence is disease or healthy. Then, they rely on examining the internal weights of the models to find the interacting SNPs. A high weight means the corresponding SNPs are contributing to predict the disease. If multiple high weight SNPs are detected, then they are considered interacted. For example, Random Forest models each node as an SNP and grows a classification tree and later examines the decision trace for interpretation~\citep{jiang2009random}; BEAM (Bayesian Epistasis Association Mapping) uses MCMC algorithm to iteratively test each marker's probability of association with disease, dependent on other markers~\citep{zhang2011bayesian}. However, most machine learning based algorithms suffer from a limited number of input sequences compared to the size of the sequence (\#SNPs). Another interesting approach is ant colony optimization algorithm~\citep{Wang2010}, which finds a refined subset of SNPs by iteratively updating a selection probability distribution. 

Although there are efficient methods to measure if a given SNPs set interact, previous works all suffer from the high computational cost of finding all possible n-combinations of SNP. For example, for a standard GWAS dataset with $10^6$ SNPs, a 2-locus exam requires $5*10^{11}$ searches, a 3-locus exam asks for $1.6*10^{17}$, a 4-locus search needs $4*10^{22}$ iterations. Hence, how to utilize these metrics to get an SNP set from a genome-scale data is the challenging part. Another challenge is that all the algorithms above assume and output fixed n-locus interactions (typically 2 or 3) where n is unknown for real biological data. We tackle these two challenges by introducing a novel model based on Reinforcement Learning to the task of epistasis detection.

\section{Method}
\subsection{Model}
\begin{figure}[]
    \centering
    \includegraphics[scale=0.7]{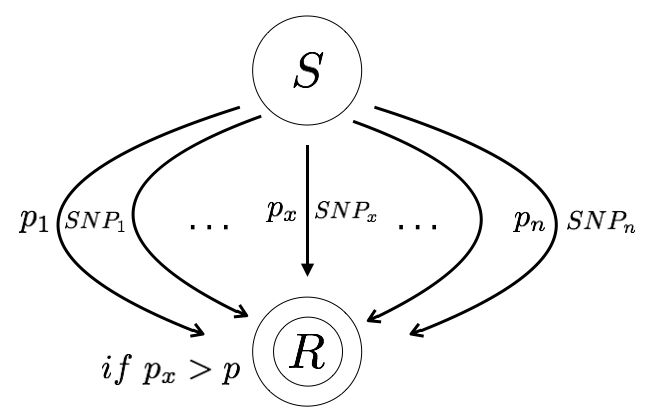}
    \caption{An illustration of our One-Step MDP Model. $S$ is genome data, and the state has $l$ actions where a probability $p_x$ is associated with the action $SNP_x$. All $SNP_x$ whose $p_x$ are larger than a threshold $p$ would be selected as an interaction set and reward $R$ is computed based on the set. Then the One-Step MDP terminates.}
    \vspace{-4mm}
\end{figure}
A typical GWAS dataset contains examples of sequences with no disease (control) and with disease (case), where both have $l$ SNPs. We denote $t_1$ and $t_2$ as the number of control and case sequences, respectively. Each SNP has three genotypes $\{aa,Aa,AA\}$, which is encoded by $\{0,1,2\}$. We want to find a set of highly interacted SNPs with the size from $2$ to $n$.

We model the epistasis process as a one-step Markov Decision Process (MDP) (Figure 1). The state $S$ is a latent representation encoded from genome data; The action space is all the SNPs, where highly interacted SNPs are selected by a probability threshold so that it poses no constraint to fix the size of interaction; the reward is efficient interaction measurements like MDR correct classification rate (CCR) and Rule Utility~\citep{yang2018multiple}. A reinforcement learning agent will learn to select SNPs that have high rewards, i.e., high interaction, by using the policy gradient method. Our approach solves the challenges mentioned above because firstly since it optimizes over iterations and chooses only a small set of actions, it is non-exhaustive, which means computationally feasible. It also utilizes the efficacy of interaction measurements like MDR CCR and Rule Utility. Second, it picks an action as long as the action passes a probability threshold, which means it can output a different size of interaction set every iteration. 

\subsection{Network}
\begin{figure}[]
    \centering
    \includegraphics[scale=0.7]{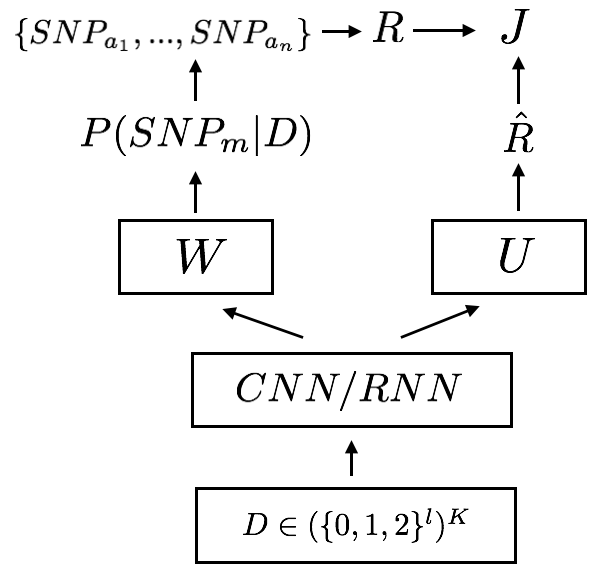}
    \caption{An illustration of our EpiRL agent. The agent first encodes D, a mini-batch of genome data, and then it predicts action-values through $W$ where a set of actions are selected and reward calculated. Along with the baseline reward, we compute the loss and iterate to learn the best actions with Reinforcement Learning.}
\end{figure}
For the input $D$, we randomly sample $K$ sequences, half from the case and the other half from the control data set (Figure 2). We then encode each sequence using the output of Convolutional Neural Network (CNN) or last hidden state of Recurrent Neural Network(RNN) to capture the spatial structure of the genome. These $K$ latent representations will be the state for our EpiRL agent. 

We then feed the state into a two-layer neural network $W$, which serves as a value function approximator. The neural network will output $l$ probabilities $P(SNP_m|D)$ for every SNP. We determine the size of interactions $n$ as the number of SNPs that have probabilities larger than $1/n$ to allow up to $n$-locus interaction. We then sample $n$ SNPs based on the probability distribution generated by the network to ensure exploration for our RL agent. This filtering forms our interaction set $I=\{SNP_{a_1}, SNP_{a_2},..,SNP_{a_n}\}$.

\subsection{Reward}
Given this SNP set $I$, we calculate the reward, which measures the interaction. Our method uses the sum of two metrics as a reward: MDR CCR and Rule Utility~\citep{yang2018multiple}. These two measures are based on MDR~\citep{motsinger2006multifactor}, which is a procedure that collapses the selected interacted data set into a four variable table. Then we perform two statistical calculations on top of this table, described as follows.

We have a genome data with size $(t_1+t_2) \times l$, where $t_1$ and $t_2$ are the number of sequences in control and case, respectively, and $l$ is the total number of SNPs. Suppose our RL agent picks $n$ actions. We then extract the genome data with these selected SNPs and form a sub data set $I$ with size $(t_1+t_2) \times n$. 

There are three genotypes for each SNP: $\{0,1,2\}$. Therefore, for $n$ SNPs, there are $3^n$ possible combinations of SNPs. We denote each combination as $\alpha_t$ where $t \in \{1,\cdots,3^n\}$. For each combination, we can count the number of control and case in $I$. We denote them $\alpha_t^1$ and $\alpha_t^2$. We assign a binary category to each combination: if $\frac{\alpha_t^2}{\alpha_t^1} < 1$, then this combination is in the low-risk group $LR$, and if $\ge 1$, it is called a high-risk group $HR$. Basically, for this specific genotype, if the number of the case exceeds the number of control, then it is high-risk, and vice versus. Now, we can construct four variables:
\begin{table}[h]
\centering
\begin{tabular}{@{}lll@{}}
\toprule
        & High Risk & Low Risk \\ \midrule
Case    & TP        & FN       \\
Control & FP        & TN      
\end{tabular}
\end{table}

where 
\begin{equation}
    TP=\sum_{t \in HR} \alpha_t^2 
\end{equation}
\begin{equation}
    FP=\sum_{t \in HR} \alpha_t^1
\end{equation}
\begin{equation}
    FN=\sum_{t \in LR} \alpha_t^2 
\end{equation}
\begin{equation}
    FP=\sum_{t \in LR} \alpha_t^1 
\end{equation}
The above equations mean that we first divide case and control sequences in the low and high-risk group and then retrieve the number of cases and controls in each group. Now, we can calculate the two measures. These two measures together are shown to be effective in measuring epistasis~\citep{yang2018multiple}. MDR CCR is the correct classification rate and Rule Utility $U$ derives from the chi-square statistics of rule relevance, which measures the interaction:
\begin{equation}
    CCR=0.5 \cdot (\frac{TP}{TP+FN}+\frac{TN}{FP+TN})
\end{equation}
\begin{equation}
    U=\frac{(R-\delta)^2}{(1+\delta)(\gamma-\delta-1)}
\end{equation}
where 
\begin{equation}
    R=\frac{FP+TN}{TP+FN}
\end{equation}
\begin{equation}
    \delta=\frac{FP}{TP}
\end{equation}
\begin{equation}
    \gamma=\frac{TP+FP+TN+FN}{TP}
\end{equation}
We sum CCR and U as our reward. Note that the calculation is fast since $n$ is usually a small number. In our preliminary study on a set with 100 SNPs, the average running time for one iteration is $0.01s$, where an iteration consists that the network predicts probabilities, calculates the reward and back-propagates the gradient.

\subsection{Training}
We train the model using REINFORCE algorithm~\citep{Williams1992}. Our objective consists of three parts: 
\begin{equation}
    J_1=(R-\hat{R})\sum_{t\in I} -\log P(t|D)
\end{equation}
\begin{equation}
    J_2=||R-\hat{R}||^2
\end{equation}
\begin{equation}
    J_3=\lambda \sum_{t \in L} p(t|D) \log P(t|D)
\end{equation}
$\hat{R}$ is a baseline reward computed by the value network $U$, a 2-layer neural network that minimizes $J_2$. $J_1$ is the advantage policy gradient. The advantage is the gap between reward and baseline, which ensures the agent to prefer actions that output rewards higher than expected. $J_3$ is the entropy regularization across all SNPs $L$ to mitigate peaky probability distribution, where $\lambda$ is the parameter to adjust the intensity of the mitigation. 

\section{Experiment}
We use simulated data from GAMETES software, which generates random, pure, strict, n-locus epistasis model~\citep{urbanowicz2012gametes}. To evaluate our method, we record SNPs set with top $K$ rewards across $C$ generated datasets. We compare this set with the ground truth labels and compute the recall $R@K=\frac{L}{C}$ where the agent gets $L$ predictions right in $C$ data set. We are also interested in the average time the agent takes to detect the right interaction.

In our preliminary study, we experiment our agent in a simulated 2-locus dataset with 600 sequences of the case and control set and with 100 SNPs. We design our data with standard genome constraint: 0.2 heritability; 0.7 prevalence; 0.2 minor allele frequency for both of 2 interacting SNPs. We minimize our objectives using the Adam optimizer~\citep{kingma2014adam} with learning rate $1e^{-3}$. 

We experimented the RL agent 50 times on the same data set. In each round of experiment, the RL agent is asked to find the interacted 2-locus SNPs under 5000 iterations. Out of the 50 trials, 34 times the agent finds the interacted SNPs under 5000 iterations. In the 34 times that the agent successfully predicts the interaction set, the average iteration is 2260.6 and the average time to find the SNPs is 22.4 s. In comparison, the exhaustive search takes 51 s. 

In the future, we will experiment on a larger dataset with various locus interactions. We will compare the recalls and the average running time with existing methods: MDR, BEAM, and Ant Optimization. At last, we will run the agent on GWAS Coronary Artery Disease (CAD) dataset since CAD is shown under epistasis effect and we will compare other study's reported epistasis on CAD with ours. 

\section{Conclusion}
Our work proposes a novel approach to model epistasis detection as a one-step MDP and introduces reinforcement learning to address this problem. We believe this will lead a new path to tackle the computational challenge in gene-gene interaction detection. 

\bibliographystyle{iclr2019_conference}
\bibliography{main}

\end{document}